%% file: main.tex
\documentclass[letterpaper, 10 pt, conference]{ieeeconf}  

\IEEEoverridecommandlockouts                              

\usepackage[font=small,labelfont=bf,compatibility=false]{caption}

\usepackage{amsmath}
\usepackage{xcolor}
\usepackage{booktabs}
\usepackage{multirow}
\usepackage{graphicx}
\usepackage{tabularx}

\usepackage{blindtext}
\usepackage{cite}
\usepackage{adjustbox}
\usepackage{soul}
\usepackage{hyperref}
\usepackage{relsize}
\usepackage{amssymb}
\usepackage{pifont}     

\def\BibTeX{{\rm B\kern-.05em{\sc i\kern-.025em b}\kern-.08em
    T\kern-.1667em\lower.7ex\hbox{E}\kern-.125emX}}

\newcommand{\cmark}{\ding{51}}%
\title{\LARGE \bf Potential Fields as Scene Affordance for \\Behavior Change-Based Visual Risk Object Identification}

\author{Pang-Yuan Pao, Shu-Wei Lu, Ze-Yan Lu, Yi-Ting Chen$^{\ddagger}$ \\
Department of Computer Science\\
National Yang Ming Chiao Tung University \\
\thanks{$^{\ddagger}$ Corresponding Author.}
}

\begin{document}

\maketitle
\thispagestyle{empty}
\pagestyle{empty}

\input{Content/0-abstract}

\section{Introduction}
\input{Content/1-introduction}

\section{Related Work}
\input{Content/2-relatedwork}
\section{Methodology}
\input{Content/3-method}

\section{Experimental Setting}
\input{Content/4-experiment}

\section{Experimental Results and Discussions}
\input{Content/5-result}

\section{Conclusion}
\input{Content/6-conclusion}

\clearpage
\bibliographystyle{IEEEtran}
\bibliography{main.bbl}

\end{document}

%% file: Content/0-abstract.tex
\begin{abstract}
We study behavior change-based visual risk object identification (Visual-ROI),  a critical framework designed to detect potential hazards for intelligent driving systems.
Existing methods often show significant limitations in spatial accuracy and temporal consistency, stemming from an incomplete understanding of scene affordance.
For example, these methods frequently misidentify vehicles that do not impact the ego vehicle as risk objects.
Furthermore, existing behavior change-based methods are inefficient because they implement causal inference in the perspective image space.
We propose a new framework with a Bird’s Eye View (BEV) representation to overcome the above challenges.
Specifically, we utilize potential fields as scene affordance, involving repulsive forces derived from road infrastructure and traffic participants, along with attractive forces sourced from target destinations.
In this work, we compute potential fields by assigning different energy levels according to the semantic labels obtained from BEV semantic segmentation.
We conduct thorough experiments and ablation studies, comparing the proposed method with various state-of-the-art algorithms on both synthetic and real-world datasets.
Our results show a notable increase in spatial accuracy and temporal consistency, with enhancements of 20.3\% and 11.6\% on the RiskBench dataset, respectively.
Additionally, we can improve computational efficiency by 88\%.
We achieve improvements of 5.4\% in spatial accuracy and 7.2\% in temporal consistency on the nuScenes dataset.
For more qualitative results, please visit our project webpage: \textcolor[RGB]{255,0,128}{\href{https://hcis-lab.github.io/PF-BCP/}{project webpage}}.

\end{abstract}

%% file: Content/1-introduction.tex
\label{sec:Introduction}
Intelligent driving systems such as advanced driver assistance systems attract significant attention in academia and industry, aiming to provide immediate alerts to reduce the number of road accidents.
One of the indispensable technologies is visual risk object identification \textbf{(Visual-ROI)}, which involves localizing potential hazards and estimating the corresponding risk scores.
The community has explored a wide range of approaches, including collision prediction~\cite{chan2016anticipating,suzuki2018anticipating,Herzig_iccvw2019,fang2019dada,You_eccv2020},
trajectory prediction and collision checking~\cite{TraPHic_CVPR2019,titan_CVPR2020,Neumann_CVPR2021},
object importance estimation~\cite{Spain_importance_eccv2008,Bar2017,zeng2017agent,Gao_goal_icra2019,zhang2020interaction},
human gaze prediction~\cite{Alletto_Dreye_cvprw2016,xia2018predicting,Xia_attention_wacv2020,Pal2021,Baee_MEDIRL_iccv2021},
and behavior change-based prediction~\cite{li2020gcn,li2020make,gupta2023object,li2023TPAMI,corl_semantic_region_2022}.

\begin{figure}[t!]
    \centering
    \vspace{-1.5em}
    \includegraphics[width=1.0\columnwidth,clip]{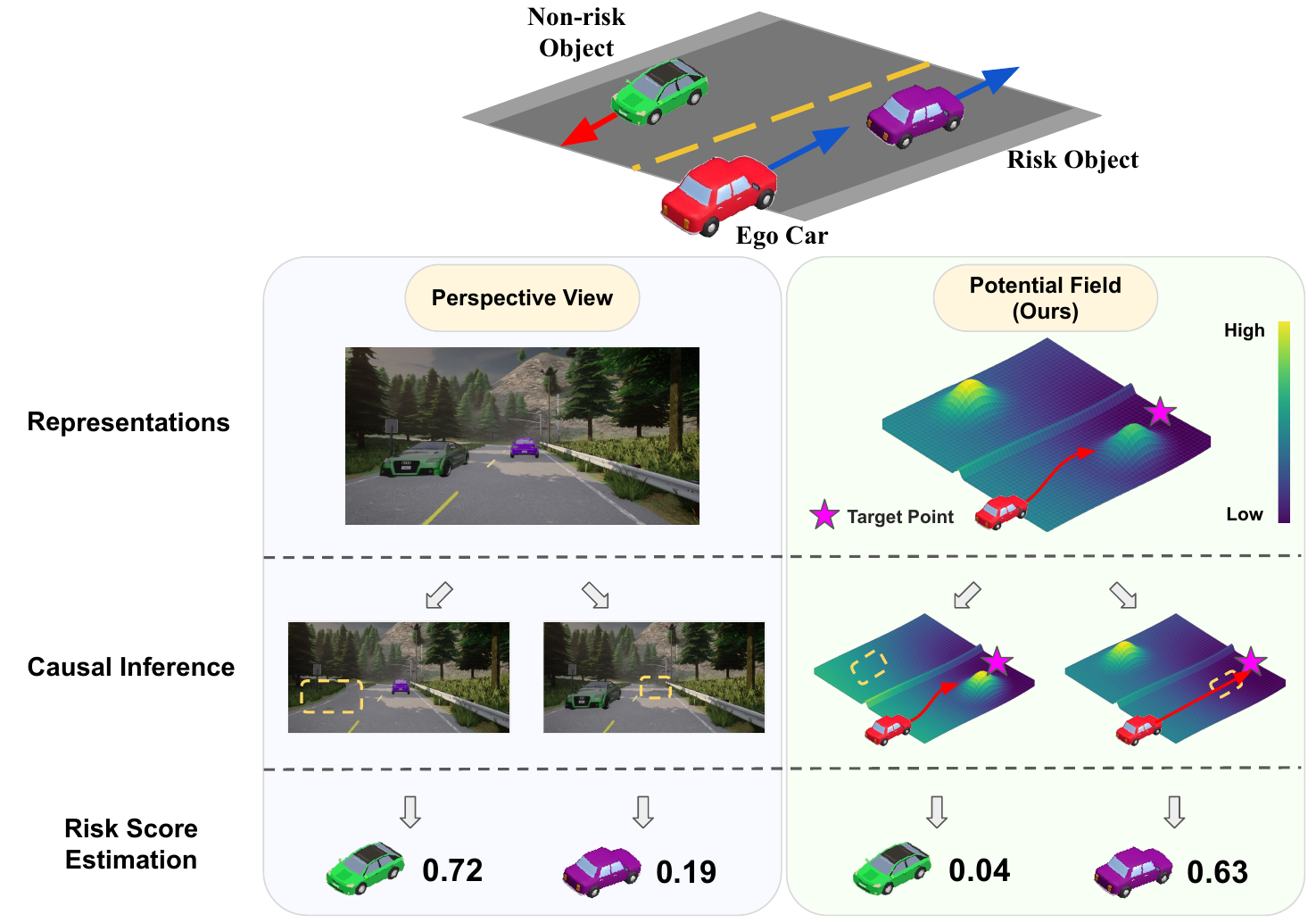} 
    \caption{\textbf{The comparison between existing behavior change-based Visual-ROI and the proposed framework.}
    Existing behavior change-based Visual-ROI algorithms~\cite{li2020make,li2023TPAMI} identify risk objects by formulating the task as a cause-effect problem.
    We identify two challenges in the existing works.
    %
    First, the approach lacks an understanding of scene affordance.
    Second, causal inference in perspective image space is time-consuming.
    Therefore, we propose \textit{potential field} as a unified representation in the bird's eye view space to address the two challenges.}
    
    \label{fig:teaser}
\vspace{-1.5em}
\end{figure}

This study examines the challenges of applying causal inference to identify risk objects within the behavior change-based prediction paradigm~\cite{li2020make,Agarwal_neuripsw_2022,li2023TPAMI,gupta2023object}.
In this paradigm, risk objects are defined as object tracklets that influence driver behavior or ego vehicle behavior\footnote{We use the terms “driver behavior” and “ego vehicle behavior” interchangeably, as the driver’s actions directly impact the ego vehicle's behavior.}.
While we observe significant progress in this direction, existing approaches often exhibit significant shortcomings in terms of spatial accuracy and temporal consistency~\cite{kung2024riskbench} and high computational complexity.
Specifically, they frequently misidentify vehicles that do not affect the ego vehicle’s behavior as risk objects, such as vehicles in the opposite lane, resulting in numerous false alarms.
The root cause of these misidentifications is that the models lack an understanding of scene affordance, i.e., what actions does a given scene afford an agent?
In addition, causal inference in perspective-view images is time-consuming, as it involves sequentially removing objects, inpainting, and feature extraction, making it impractical for real-world deployment.

To address these challenges, we introduce a unified framework that incorporates scene affordance, operationalized via potential fields~\cite{Khatib1985}, to address the aforementioned limitations. 
Potential fields model scene affordance via repulsive and attractive forces, as depicted in Fig.~\ref{fig:teaser}.
Road markings (e.g., solid white lines) and traffic participants (e.g., vehicles and pedestrians) are modeled with high repulsive forces.
A target destination is modeled with low attractive forces, guiding ego vehicles toward the destination.
We compute potential fields using the outputs of bird’s eye view semantic segmentation models (BEV-SEG), such as those from~\cite{wu2021cvt}. 
Moreover, we can significantly reduce inference time by eliminating the need for image inpainting and feature extraction in the perspective image space by conducting causal inference in the BEV space.

We demonstrate the effectiveness of enhancing spatial accuracy and temporal consistency, as well as improving efficiency, on the \textbf{RiskBench} (synthetic)~\cite{kung2024riskbench} and \textbf{nuScenes} (real-world)~\cite{nuscenes} datasets. 
Additionally, we perform extensive ablation studies to validate our design choices and carefully assess the significance of the various components in our methodology.
Furthermore, we present qualitative evidence showing that the use of potential fields effectively identifies risk objects. 

\vspace{1mm}
Our contributions are summarized as follows:
\begin{itemize}

\item We propose using potential fields to model scene affordance, addressing spatial inaccuracy and temporal inconsistency and inefficiency observed in behavior change-based visual risk object identification. 

\item We conduct extensive evaluations and ablation studies on the RiskBench and nuScenes datasets to demonstrate the effectiveness and justify the proposed method.

\item We present qualitative evidence showing that the proposed framework effectively identifies objects that influence ego vehicle behavior in both synthetic and real-world datasets.

\end{itemize}

%% file: Content/2-relatedwork.tex
\subsection{Visual Risk Object Identification}

Visual risk object identification (Visual-ROI)~\cite{Spain_importance_eccv2008,Alletto_Dreye_cvprw2016,chan2016anticipating,Bar2017,zeng2017agent,suzuki2018anticipating,xia2018predicting,Tawari2018,Herzig_iccvw2019,Gao_goal_icra2019,fang2019dada,TraPHic_CVPR2019,li2020make,li2020gcn,Xia_attention_wacv2020,zhang2020interaction,You_eccv2020,titan_CVPR2020,Neumann_CVPR2021,Pal2021,Baee_MEDIRL_iccv2021,gupta2023object,li2022important,li2023TPAMI} is an essential technology for the development of intelligent driving systems.
The field can be categorized into the following four paradigms.
First, predicting objects that potentially involve in collision is defined as risk object~\cite{chan2016anticipating,suzuki2018anticipating,Herzig_iccvw2019,fang2019dada,You_eccv2020, TraPHic_CVPR2019,titan_CVPR2020,Neumann_CVPR2021}. 
Second, risk objects are defined based on human annotators' subjective assessment~\cite{Spain_importance_eccv2008,Bar2017,zeng2017agent,Gao_goal_icra2019,zhang2020interaction,Sachdeva_2024_WACV}.
Third, objects focused by human gaze are defined as risk objects~\cite{Alletto_Dreye_cvprw2016,xia2018predicting,Xia_attention_wacv2020,Pal2021,Baee_MEDIRL_iccv2021}.
Forth, objects influencing either the driver's or ego vehicle's behavior are defined as risk objects~\cite{li2020make,Agarwal_neuripsw_2022,gupta2023object,li2023TPAMI,li2020gcn}.

A common challenge across these paradigms is the extraction of task-relevant information from high-dimensional visual inputs to enhance risk object identification.
Researchers have explored methods such as computing object locations, semantic layouts, and depth to better capture the context of traffic scenes.
Additionally, intents of ego vehicles~\cite{Gao_goal_icra2019,corl_semantic_region_2022} and other traffic participants~\cite{Gao_goal_icra2019,titan_CVPR2020} have been studied to provide further insights for risk object identification.
More recently, the community has also investigated using language to guide the extraction of task-relevant information from visual inputs~\cite{kim_grounding_cvpr2019,Malla_2023_WACV,Sachdeva_2024_WACV}. 
In this work, we look into scene affordance, a type of task-relevant information essential for understanding the potential actions and interactions within a traffic environment. 
%
%
By modeling these affordances, we aim to bridge the gap between raw visual inputs and higher-level reasoning, i.e., risk object identification.

\subsection{Potential Fields}
Potential fields are a fundamental technique in robotics, widely used for trajectory prediction, risk assessment, motion planning, and path planning~\cite{Khatib1985,Barraquand1991,Hwang1992,REIF1999171,Althoff_srs_iv2008,Rehmatullah2015,Rasekhipour2017,gao2019pf,Huang2020,Lu2020,Yi2020,KOLEKAR2020103196,CKolekar2020DRF,chen2022complexity,lin2023potential}.
In recent years, researchers have begun exploring the integration of machine learning with the concept of potential fields for traffic scene applications.
Su et al.~\cite{Su2019PotentialFI} propose a learning algorithm to extract three different potential fields that model environmental structure, target object historical motion, and social interaction from birds’-eye-view (BEV) images.
Recently, the community has also explored other field representations.
For instance, Mahjourian et al.,~\cite{occupancy_flow_2022} introduce occupancy flow fields, learned from sparse environments and agents' states, for motion forecasting of multiple agents.
In the context of planning tasks, variants of potential fields have been studied to capture the motion of traffic participants and the semantic structure of roads~\cite{Chitta_neat_2021,Casas_mp3_2021,hu_stp3_2022}.

%

In this work, we extend the use of potential fields as a representation of scene affordance, such as solid road markings do not afford crossing actions.
To the best of our knowledge, this is the first study to explore the use of potential fields as a representation of scene affordance in the context of Visual-ROI.
In~\cite{Chitta_neat_2021,hu_stp3_2022}, they use BEV semantic segmentation as a representation for downstream planning. 
Instead, we derive potential fields from perspective images by assigning different energy levels to the corresponding semantic labels. 
Our approach complements existing methods that model the motion of traffic participants via motion fields, and we plan to explore this synergy in our future work.

\begin{figure*}[t!]
    \centering
    \includegraphics[width=1.0\textwidth]{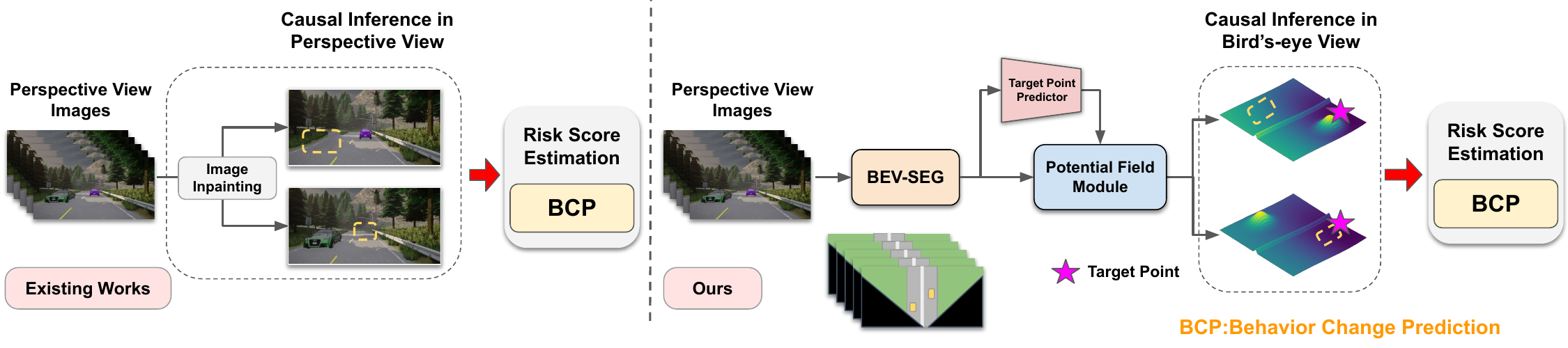}
    \caption{\textbf{Overview of Our Framework.}
    The figure compares behavior change-based Visual-ROI methods.
    On the left, existing works~\cite{li2020make,li2023TPAMI} conduct causal inference via image inpainting in perspective-view images and estimate risk scores through behavior change prediction.
    However, inpainting in perspective-view images is time-consuming because we must re-compute the corresponding image features when removing an object tracklet. 
    Moreover, the existing works do not model scene affordance, resulting in inferior spatial accuracy and temporal consistency.
    In contrast, our proposed method (right) conducts causal inference in the bird's-eye view, enabling a parallel object removal process and using potential field as a new representation of scene affordance, providing rich information for reasoning risk objects. 
    }
    \label{fig:framework}
    \vspace{-1.5em}
\end{figure*}

\subsection{Scene Affordance}

The concept of functional visual understanding, which examines the potential actions a scene offers to an agent, originates from the work of Gestalt psychologists in the 1930s~\cite{koffka1935}. 
This idea is formalized by J.J. Gibson~\cite{gibson1979}, who introduced the term "affordance."
Numerous studies have been explored, such as affordance prediction~\cite{Gupta2011scenegeo,Grabner2011Chair,Delaitre2012SceneSF,Fouhey2012peoplewatching,Jiang2013hallucinatedhumans,Fouhey2015InDO,chuang2018LAP,wang2018bingewatching,li2019puttinghumans,brooks2022hallucinatingpose} and the modeling of human-object affordances~\cite{Yao2010modelingmutualcontext,koppula2013learninghuHA,Zhu2014reasoningobjectaffordance,gkioxari2018detectingrecognizingHOI,cao2021reconstructinghandobject,Kulal_human_affordance_2023}.
In the context of intelligent driving systems, research has explored incorporating affordances into the decision-making processes of the ego vehicle.
For example, Chen et al.~\cite{DeepDriving2015} propose mapping images into key affordance factors, such as the distance to other traffic participants and road markings.
Another approach focuses on directly predicting semantic labels in the perspective view as scene affordance, such as identifying drivable areas~\cite{bdd100k}.

We exploit the concept of repulsive and attractive forces within potential fields~\cite{Khatib1985} to represent scene affordance.
A repulsive force guides an ego vehicle away from potential hazards, while an attractive force directs it toward its destination.
This dual mechanism of potential fields equips the ego vehicle with a comprehensive scene affordance.
We show that the representation can improve the performance to identify risk objects and the efficiency of causal inference due to its nature in the BEV space.

%% file: Content/3-method.tex
\label{sec:method}

Our objective is to design a framework that leverages potential fields as scene affordances to improve the spatial accuracy and temporal consistency and computational efficiency of behavior change-based Visual-ROI.
The structure of the framework is shown in Fig.~\ref{fig:framework}.
It is composed of four key components: Bird's Eye View Semantic Segmentation (BEV-SEG), target point prediction, potential field rendering, and a behavior change-based Visual-ROI predictor.
The detailed implementation can be found in our \textcolor[RGB]{255,0,128}{\href{https://github.com/HCIS-Lab/PF-BCP}{GitHub repository}}.


\subsection{BEV Semantic Segmentation}

We choose CVT~\cite{wu2021cvt} as our BEV encoder because of its lightweight structure.
CVT utilizes a transformer-like architecture that takes BEV embeddings as queries and image features as keys and values.
%
%
The input image resolution is 256$\times$640. The output BEV-SEG resolution is 100$\times$200. 
This corresponds to a [0m, 50m]$\times$[-50m, 50m] region centered around the ego vehicle.
%
%

\subsection{Target Point Prediction}

We adopt the architecture designed in AIM-BEV~\cite{Hanselmann2022ECCV} for our target point predictor.
The predictor receives a sequence of predicted BEV-SEG with a length of $N$. 
In our implementation, we set $N$ to 5, allowing the model to access the previous observations.
The sequence of predicted BEV-SEGs are processed by encoders~\cite{Hanselmann2022ECCV}.
The output features are concatenated and flattened into a one-dimensional embedding.
It is then fed into an LSTM, followed by fully connected layers for target point prediction.
%
%
%
%
%
%
The final output is a 2-dimensional representation of a target point $T_p$ in the BEV space, serving as an input for computing attractive force in potential field.

\subsection{Potential Field Rendering}

We adopt the method proposed in~\cite{Khatib1985} to render potential fields.
\\
\textbf{Repulsive Force:}
Repulsive forces are calculated based on the distance of ego vehicles to obstacles, which aims to prevent collisions and guarantee safe navigation.
For a specific point $p$ on a predicted BEV-SEG $I$, we model its repulsive force as \( F_r = \max(K_r/(\textrm{ED}(p,q))^2) \ \forall q \in I \),
where the term ED is the Euclidean distance between $p$ and $q$.
We empirically set the repulsive constants $K_r$ to be $400$ for road lines, $1000$ for dynamic objects (i.e., vehicles and pedestrians), and 0 for other objects.
\\
\textbf{Attractive Force:}
Attractive force enables the computation of a feasible path toward a desired goal location.
For a given specific point $p$ and target point $T_p$ on a predicted BEV-SEG $I$,  attractive force is formulated as \(F_a = K_a *\textrm{ED}(T_p, p)\).
We set $K_a$ to be 0.75 for the attractive constant.
Ego vehicles are afforded to act from high energy (its current location) to low energy (a target location).

%
\noindent\textbf{Potential Field:}
The complete potential field $F$ can be calculated as \(F = F_a+F_r\).
We use this model to capture scene affordance, i.e., what actions does a given scene afford an agent?
Repulsive forces informs an agent to avoid obstacles, and attractive forces guide an agent to act toward a goal location.

\subsection{Behavior Change-based Visual Risk Object Identification}
\label{sec:method-PF+BCP}
In the behavior change-based Visual-ROI paradigm~\cite{li2020make}, risk objects are defined as object tracklets that influence ego vehicle behavior.
Acoording to~\cite{li2020make}, a behavior change predictor is first trained to determine whether the ego vehicle’s behavior is influenced. 
When the predicted response is \textit{Stop}, a causal inference approach is employed to estimate the risk score of each object.
Specifically, one object tracklet is removed at a time, and inpainting is used to fill the removed area in each frame, simulating a scene without that object. 
The behavior change predictor is then reapplied to assess whether the ego vehicle’s behavior remains influenced.
The risk score of the removed tracklet is defined as the difference in predicted scores between the original scene and the counterfactual scene. 
The object that results in the highest response change is identified as the risk object.

In this work, we make the following two changes to~\cite{li2020make}, referred to as \textbf{PF+BCP}. First, we conduct causal inference in the BEV instead of the perspective view. 
Precisely, given a rendered potential field and an object tracklet, we remove the corresponding repulsive force to simulate a potential field without the tracklet.
Second, we train a behavior change predictor using the rendered potential fields as input. 
%
%
%
%
%
%

%% file: Content/4-experiment.tex
\label{sec:experiment}

\subsection{Datasets}
We conduct experiments on the \textbf{RiskBench} dataset~\cite{kung2024riskbench} and the \textbf{nuScenes} dataset~\cite{nuscenes}. 
\textbf{RiskBench} is collected in CARLA~\cite{Dosovitskiy17carla}, which is the largest scenario-based benchmark for risk object identification.
In addition, we use the \textbf{nuScenes} dataset as the testbed to evaluate the effectiveness of different Visual-ROI in the real world.
However, \textbf{nuScenes} dataset does not have the annotations of risk objects.
We label risk objects manually according to the protocol described in~\cite{li2020make}. 
%
%
For the \textbf{RiskBench} and \textbf{nuScenes} datasets, we define ground truth target points as the trajectory logs recorded 3 seconds into the future from each given a time step.

\subsection{Baselines}

We evaluate the following nine Visual-ROI baselines.

%
%
\noindent\textbf{FF~\cite{HuCVPR2021}:}
FF constructs a new self-supervised learning task called Freespace Forecasting (FF).
We use the predicted occupied score to represent the risk score of an object.

\noindent\textbf{DSA~\cite{chan2016anticipating}:}
Dynamic-Spatial-Attention classifies whether a collision will occur in a scene.
It uses a soft attention mechanism to assess how much each object contributes to collision prediction.
We use a soft attention score to denote the risk score of an object.

\noindent\textbf{RRL~\cite{zeng2017agent}:}
The method introduces a task that predicts which object is involved in a collision.
We use the predicted collision score as the risk score for each object.

\noindent\textbf{BP~\cite{li2020gcn}:}
Behavior Prediction (BP) employs graph attention networks to model interactions between traffic participants and the ego vehicle.
The object with the highest attention score is defined as the risk.

\noindent\textbf{BCP~\cite{li2020make}}:
BCP scores an object's riskiness by comparing the response changes, as discussed in section~\ref{sec:method-PF+BCP}.

\noindent\textbf{TP+BCP:}
We integrate $T_p$ from the target point predictor into BCP, using the same mechanism to define risk scores. Here, $T_p$ is represented as a 2-dimensional vector as input.

\noindent\textbf{BS+BCP:}
This baseline utilizes BEV-SEG as the input for BCP and uses the same mechanism to define their risk scores.

\noindent\textbf{OADE:}
Following OIECR~\cite{gupta2023object}, the risk score for an object is defined as the average displacement error (ADE) between the planned waypoints generated from the original observation and counterfactual observation (i.e., the observation with the removed object).
The planned waypoints are determined by following the gradient of rendered potential fields from high to low potential without any reversals.
The trajectory originates at the ego vehicle and, if possible, ends at the predicted target point.

\noindent\textbf{OFDE:}
In a manner similar to OADE, we compute the final displacement error (FDE) as the risk score of an object.

\subsection{Metrics}

We evaluate Visual-ROI models with two types of metrics: \textbf{Spatial Accuracy} and \textbf{Temporal Consistency}.
\textbf{Spatial Accuracy} involves Optimal F1 Score (OT-F1) and Optimal F1 Score in T Seconds (OT-F1-T).
%
\textbf{Temporal Consistency} consists of Progressive Increasing Cost (PIC) and Weighted Multi-Object Tracking Accuracy (wMOTA).\\
%
\noindent\textbf{Optimal F1 Score (OT-F1):}
An object is a risk object if its raw risk score exceeds a certain threshold. 
The optimal threshold is selected by maximizing the F1 score through a sweeping analysis. 
This process serves as an upper-bound performance benchmark for each model.\\
%
%
\noindent\textbf{Optimal F1 Score in T Seconds (OT-F1-T):}
OT-F1-T evaluates the OT-F1 prediction outcomes during the T seconds preceding the critical point. 
A critical point is defined as the moment when the ego vehicle is both influenced by the risk object and is at its closest proximity to it~\cite{kung2024riskbench}.\\
\textbf{Progressive Increasing Cost (PIC):}
This metric is introduced in RiskBench~\cite{kung2024riskbench}. 
To address the issue of penalty weights decreasing too quickly and approaching zero, we adjust PIC as
\(\text{PIC} = -\sum^{T}_{t=1} e^{-(T-t)/T}\log(\text{F1}_{t})\).
Here, $\textrm{F1}_t$ denotes the F1 score at a specific time frame $t$, while $T$ represents the total number of frames within a scenario. 
We establish $T$ as 60, equivalent to 3 seconds. 
We scale PIC to a range between 0 and 1 for improved interpretability by aggregating the PIC values across all scenarios and normalizing the total PIC to fit within this scale. \\
\noindent\textbf{Weighted Multi-Object Tracking Accuracy (wMOTA):}
Inspired by MOT16~\cite{milan2016mot16}, we use MOTA to evaluate the temporal consistency of a Visual-ROI model.
%
To address the imbalance between positive (risky) and negative (non-risky) samples, we propose a weighted version of MOTA called wMOTA.
We denote the number of positive miss at time $t$ as $\textrm{PM}_t$. 
The number of negative miss at time $t$ as $\textrm{NM}_t$. 
The value $\textrm{PM}_t$ is defined as $w_p\cdot(\textrm{FN}_t+\textrm{IDsw}^{p}_t)$.
The value $\textrm{NM}_t$ is defined as $w_n\cdot(\textrm{FP}_t+\textrm{IDsw}^{n}_t)$.
In the above two equations, the notations $\textrm{FN}_t$ and $\textrm{FP}_t$ represent the numbers of false negatives and false positives at time $t$, respectively.
In addition, the notations $\textrm{IDsw}^p_t$ and $\textrm{IDsw}^n_t$ represent the number of identity switches for positive and negative samples, respectively, between times $t-1$ and $t$.
The parameters $w_p$ and $w_n$ are the weights assigned to positive and negative samples.
The wMOTA is defined as
\(\text{wMOTA}=1-\frac{\sum_t(\textrm{PM}_t+\textrm{NM}_t)}{\sum_t(w_p{\cdot}\textrm{GT}^{p}_t+w_n{\cdot}\textrm{GT}^{n}_t)}\),
where $\textrm{GT}^{p}_t$ and $\textrm{GT}^{n}_t$ represent the counts of ground truth positive and negative samples at time $t$, respectively.

\vspace{1em}

%% file: Content/5-result.tex
\label{sec:result}

\vspace{1em}

\subsection{Visual Risk Object Identification}

We conduct extensive experiments on the RiskBench dataset~\cite{kung2024riskbench}. We show the proposed method \textbf{PF+BCP} achieves state-of-the-art performance compared to existing Visual-ROI methods, as shown in Table~\ref{table:ROI_main_table}.
Our method has shown remarkable effectiveness, yielding a 20.3\% improvement in the F1 score over the \textbf{BCP}~\cite{li2020make} on the RiskBench dataset.
Additionally, we achieve a 2.6\% improvement in the F1 score over \textbf{BS+BCP}, highlighting the benefit of incorporating attractive and repulsive forces into the model.
Note that the improvements in both recall and precision further validate the robustness of our approach.

In terms of temporal consistency, \textbf{PF+BCP} shows an improvement in wMOTA by 11.6\% compared to \textbf{BCP} and 2.3\% compared to \textbf{BS+BCP}.
The improvements in PIC further validate that the proposed method enhances spatial accuracy and temporal consistency over existing models.
A similar trend is also observed in the OT-F1-T results.
As shown in Table~\ref{table:OT-F1-T}, \textbf{PF+BCP} improves the OT-F1-T score by 13–15\% compared to \textbf{BCP}, demonstrating that the proposed method provides stable predictions near the critical frames.

\begin{table}[t!]
    \footnotesize
    \centering
    \setlength{\tabcolsep}{0.73mm}
    \renewcommand{\arraystretch}{1.0}
    \captionof{table}{\textbf{Evaluation on the RiskBench.}
    The notations \textbf{P} and \textbf{R} denote precision and recall, respectively.
    All units are presented as percentages, except for inference time.
    The best results are highlighted in bold, and the second are underlined.}
    \begin{tabular}
        {@{}l c c c | c c |@{} c@{}}
        \toprule
        & \multicolumn{3}{c}{Spatial Accuracy}
        & \multicolumn{2}{c}{Temporal Consistency}
        & \multicolumn{1}{c}{Inference Time} \\
        \cmidrule(lr){2-4} \cmidrule(lr){5-6} \cmidrule(lr){7-7}
        & OT-P$\mathsmaller{\uparrow}$
        & OT-R$\mathsmaller{\uparrow}$
        & OT-F1$\mathsmaller{\uparrow}$
        & ~~~PIC$\mathsmaller{\downarrow}$~~~\ 
        & wMOTA$\mathsmaller{\uparrow}$
        & Avg (sec)$\mathsmaller{\downarrow}$ \\
        \midrule
        FF~\cite{HuCVPR2021} 
        & 22.2 & 27.9 & 24.7 & 39.3 & 55.0 & \textbf{0.027} 
        \\
        DSA~\cite{chan2016anticipating}
        & 54.7 & 19.7 & 29.0 & 29.8 & 53.3 & 0.269
        \\
        RRL~\cite{zeng2017agent}
        & 49.4 & 15.4 & 23.5 & 28.9 & 52.3 & 0.280
        \\
        BP~\cite{li2020gcn}
        & 24.2 & 35.1 & 28.7 & 39.0 & 57.5 & 0.119
        \\
        BCP ~\cite{li2020make}
        & 38.6 & 43.7 & 41.0 & 29.3 & 63.2 & 0.431
        \\
        \midrule
        TP+BCP
        & 47.4 & 51.7 & 49.5 & 28.0 & 67.2 & 0.437
        \\
        BS+BCP
        & \underline{56.8} & \underline{60.7} & \underline{58.7} & \underline{24.0} & \underline{72.5} & \underline{0.049}
        \\
        \midrule
        OFDE
        & 50.8 & 56.7 & 53.6 & 26.7 & 65.4 & 0.062
        \\
        OADE
        & 52.7 & 57.9 & 55.2 & 25.7  & 66.9 & 0.061
        \\
        PF+BCP
        & \textbf{60.2} & \textbf{62.4} & \textbf{61.3} & \textbf{23.0} & \textbf{74.8} & \underline{0.049}
        \\
        \bottomrule
    \end{tabular}
    \label{table:ROI_main_table}
    \vspace{-1em}
\end{table}

Moreover, we demonstrate that employing BEV representations (\textbf{BS+BCP, PF+BCP}) for causal inference significantly enhances inference speed, achieving an 88\% improvement compared to \textbf{BCP}, which employs perspective-view images.
It is worth noting that the overall performance of \textbf{OADE} and \textbf{OFDE} is slightly inferior compared to \textbf{PF+BCP}.
This discrepancy is likely due to the strong reliance of \textbf{OADE} and \textbf{OFDE} on upstream BEV-SEG quality. 
Pixel-level noise in the BEV-SEG can introduce significant errors in the potential field, impacting overall performance.

\begin{table}[!h]
    \centering
    \setlength{\tabcolsep}{2em}
    \renewcommand{\arraystretch}{1.0}
    \captionof{table}{\textbf{OT-F1-T.}
    OT-F1 Performance during the T seconds preceding the critical frame. }
    \begin{tabular}
    {@{} l @{\;} c @{\;} c @{\;} c @{\;} c @{\;}}
        \toprule
        \multicolumn{1}{@{}l}{Method}& 
        \multicolumn{1}{c}{1s $\mathsmaller{\uparrow}$} & 
        \multicolumn{1}{c}{2s $\mathsmaller{\uparrow}$} & 
        \multicolumn{1}{c}{3s $\mathsmaller{\uparrow}$} &
        \multicolumn{1}{c}{Overall $\mathsmaller{\uparrow}$}
        \\ 
        \midrule
        FF~\cite{HuCVPR2021} 
        & 28.7 & 24.4 & 21.5 & 24.7
        \\
        DSA~\cite{chan2016anticipating} 
        & 36.8 & 31.6 & 29.7 & 29.0
        \\
        RRL~\cite{zeng2017agent} 
        & 35.0 & 32.2 & 31.9 & 23.5
        \\
        BP~\cite{li2020gcn} 
        & 33.8 & 32.8 & 30.8 & 28.7
        \\
        BCP~\cite{li2020make} 
        & 49.3 & 47.2 & 44.2 & 41.0
        \\
        \midrule
        TP+BCP
        & 52.8 & 49.8 & 46.9 & 49.5
        \\
        BS+BCP
        & 60.7 & 58.8 & 56.5 & 58.7
        \\
        \midrule
        OFDE
        & 56.4 & 53.0 & 50.3 & 53.6
        \\
        OADE
        & 57.9 & 55.0 & 52.7 & 55.2
        \\
        PF+BCP
        & \textbf{62.5} & \textbf{61.0} & \textbf{59.3} & \textbf{61.3}
        \\
        \bottomrule
    \end{tabular}
    \label{table:OT-F1-T}
\end{table}

\subsection{Scenario-based Analysis}
We perform a scenario-based analysis to assess model performance in various traffic conditions.
As shown in Table~\ref{table:oncominglane}, \textbf{PF+BCP} demonstrates fewer false positives compared to all baselines in the opposite-lane scenario. 
This strong performance is due to its focus on objects that may influence the ego vehicle’s progress toward its target (attractive force). 
Moreover, the modeling of repulsive force helps identify non-afforded actions toward objects in the opposite lane.

\begin{figure}[t!]
    \begin{minipage}[t!]{0.22\textwidth}
    \scriptsize 
    \centering
    \setlength{\tabcolsep}{2.3mm}
    \renewcommand{\arraystretch}{1.0}
    \captionof{table}{\textbf{Opposite-Lane Situation.}
    \textbf{FP} and \textbf{TN} refer to false positives 
    and true negatives, respectively.
    }
    \begin{tabular}
        {@{}l @{\;} c @{\;} c @{\;} c @{\;} @{}c @{}}
        \toprule
        \multicolumn{1}{@{}l}{Method} &
        \multicolumn{1}{c}{~FP$\mathsmaller{\downarrow}$}  & 
        \multicolumn{1}{c}{~TN$\mathsmaller{\uparrow}$}  & 
        \multicolumn{1}{@{}c@{}}{Acc(\%)$\mathsmaller{\uparrow}$}
        \\ 
        \midrule
        BP~\cite{li2020gcn}
        & 1,861 & 26,486 & 93.4
        \\
        BCP~\cite{li2020make}
        & 1,951 & 26,393 & 93.1
        \\
        \midrule
        TP+BCP
        & 1,105 & 27,242 & 96.1
        \\
        BS+BCP
        & 1,323 & 27,024 & 95.3
        \\
        \midrule
        OFDE
        & 1,432 & 26,915 & 94.9
        \\
        OADE
        & 1,190 & 27,157 & 95.8
        \\
        PF+BCP
        & \textbf{694} &\textbf{27,653} & \textbf{97.6}
        \\
        \bottomrule
    \end{tabular}
    \label{table:oncominglane}
    \end{minipage}
    \hfill
    \begin{minipage}[t!]{0.25\textwidth}
    \scriptsize 
    \centering
    \setlength{\tabcolsep}{0.7mm}
    \renewcommand{\arraystretch}{1.0}
    \captionof{table}
    {\textbf{Actor-Based Analysis.}
    \textbf{Clst} and \textbf{Peds} refer to closest actors and pedestrians, respectively. The metric is OT-F1.
    }
    \begin{tabular}
        {@{} l c c c c }
        \toprule
        \multicolumn{1}{@{}l}{Method} & 
        \multicolumn{1}{c}{Overall$\mathsmaller{\uparrow}$}  & 
        \multicolumn{1}{c}{Peds$\mathsmaller{\uparrow}$}  & 
        \multicolumn{1}{c}{Trucks$\mathsmaller{\uparrow}$} & 
        \multicolumn{1}{c}{Clst$\mathsmaller{\uparrow}$} 
        \\ 
         \midrule
         BP~\cite{li2020gcn} 
         & 28.7 & 26.6 & 37.5 & 37.2
         \\
         BCP~\cite{li2020make} 
         & 41.0 & 33.7 & 52.0 & 48.3
         \\
         \midrule
         TP+BCP 
         & 49.5 & 43.2 & 57.8 & 56.3
         \\
         BS+BCP 
         & 58.7 & 44.4 & 71.6 & 62.3
         \\
         \midrule
         OFDE
         & 53.6 & 39.7 & 67.2 & 58.4
         \\
         OADE
         & 55.2 & 42.8 & 68.3 & 59.6
         \\
         PF+BCP
         & \textbf{61.3} & \textbf{46.9} & \textbf{73.5} & \textbf{64.7}
         \\
        \bottomrule
    \end{tabular}
    \label{table:actor-based}
    \end{minipage}
    \vspace{-1em}
\end{figure}

\textbf{PF+BCP} significantly outperforms other methods in scenarios involving the various actors, as shown in Table~\ref{table:actor-based}. 
The results from the \textbf{Closest Actors} demonstrate that the \textbf{PF+BCP} method effectively identifies risk traffic participants in the vicinity of the ego vehicle.
The \textbf{Truck} scenario demonstrates superior performance across all methods, likely because BEV semantic segmentation can detect large objects.
In contrast, the \textbf{Pedestrians} scenarios show less satisfactory results, as BEV-SEG struggles to accurately capture pedestrian features.
This limitation leads to errors in the causal inference stage and impacts the performance of Visual-ROI.
We find that small objects, such as pedestrians, pose a unique challenge, highlighting the need for further advancement in BEV semantic segmentation.
Our scenario-based analysis demonstrates the versatility and effectiveness of our approach across various driving conditions, while also identifying the need for improvement in small object detection.

\begin{figure*}[t!]
    \centering
    \includegraphics[width=1.0\textwidth]{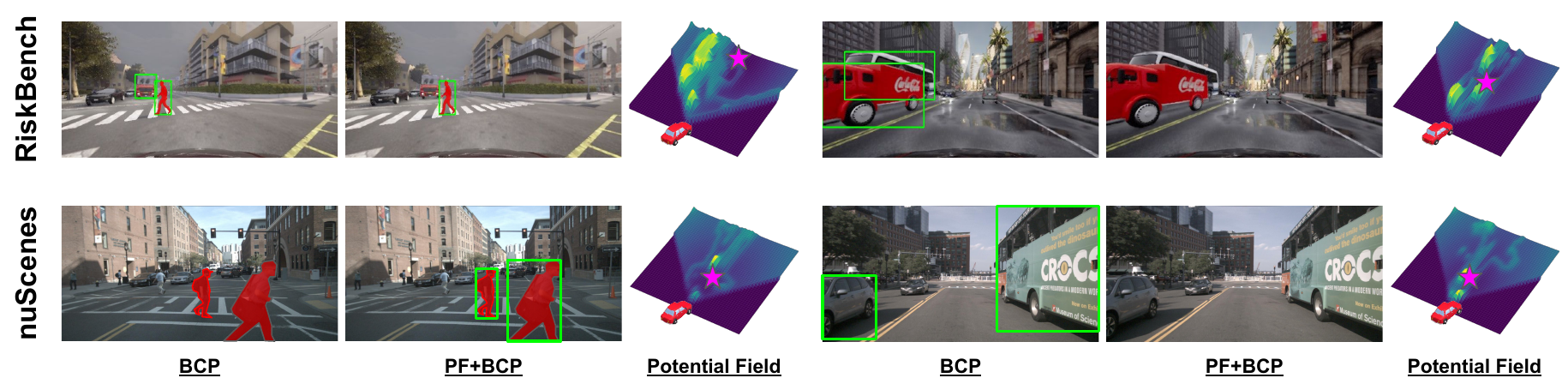}
    \vspace{-1.5em}
    \caption{
    Visualization of ROI results on sampled scenarios selected from the \textbf{RiskBench} and \textbf{nuScenes} dataset.
    All detected risk objects are shown with green bounding boxes, while ground truth risks are masked in red.
    Target points are marked with a purple star.}
    \label{fig:qualitative_ROI}
    \vspace{-2ex}
\end{figure*}

\begin{table}[h!]
    \centering
    \small
    \setlength{\tabcolsep}{2.9mm}
    \renewcommand{\arraystretch}{1.0}
    \captionof{table}{\textbf{Ablation Studies.} 
    The table shows the performance of different combinations of modules, including BEV-SEG (BS), repulsive force $F_r$, and attractive force $F_a$, when integrated with BCP.
    Note that in \textbf{ID 2} and \textbf{ID 3}, the notation $F_a$ represents a 2-dimensional vector $T_p$ corresponding to the target point.
    }
    \begin{tabular}
    {@{}l@{\;}@{\;} | @{} c @{} c @{\;} c @{\;} | @{\;} c @{\;} c @{\;} c @{\;}}
        \toprule
        \multicolumn{1}{@{}l}{ID}  &
        \multicolumn{1}{c}{~BS~}  &
        \multicolumn{1}{c}{${F_r}$}  &
        \multicolumn{1}{c}{${F_a}$}  &
        \multicolumn{1}{c}{OT-F1 $\uparrow$} &
        \multicolumn{1}{c}{PIC $\downarrow$} &
        \multicolumn{1}{c}{wMOTA $\uparrow$}
        \\
        
        
        \midrule
        1 & & & & 41.0 & 29.3 & 63.2
        \\
        2 & & & \cmark & 49.5 & 28.0 & 67.2
        \\
        3 & \cmark & & \cmark & 58.7 & 24.0 & 72.5
        \\
        4 & & \cmark & & 59.0 & 24.4 & 72.9
        \\
        5 & & \cmark & \cmark & \textbf{61.3} & \textbf{23.0} & \textbf{74.8}
        \\
        \bottomrule
    \end{tabular}
    \label{tab:ablation}
    \vspace{-1em}
\end{table}

\subsection{Ablation Studies}

We conduct extensive ablation studies to justify our design choices built upon BCP \textbf{(ID 1)}~\cite{li2020make}. 
The results are presented in Table~\ref{tab:ablation}.
Adding a target point predictor to BCP \textbf{(ID 2)} improves OT-F1 by 8\%, highlighting the importance of modeling ego vehicle’s intention. 
Incorporating BEV semantic segmentation \textbf{(ID 3)} further enhances OT-F1 by nearly 10\%, demonstrating that rich semantic context aids in risk object identification.
In \textbf{ID 4}, we assess the importance of the repulsive force without attractive force and find its performance comparable to \textbf{ID 3}.
We conjecture that the lack of attractive force may not be fully aware of the risk object toward the target location, increasing the number of misidentifications. 
Finally, combining all the designed modules \textbf{(ID 5)} yields the best results, i.e., an 20\% improvement over BCP.

\subsection{Real-world Evaluation}
We conduct a real-world evaluation on the nuScenes~\cite{nuscenes} dataset. 
As shown in Table~\ref{table:nuscenes}, \textbf{PF+BCP} outperforms the baselines \textbf{BP}, \textbf{BCP} and \textbf{BS+BCP} in both spatial accuracy and temporal consistency metrics. 
Specifically, compared to \textbf{BP} and \textbf{BCP}, \textbf{PF+BCP} improves OT-F1 by 29\% and 5.4\%, reduces PIC by 10.1\% and 6.3\%, and enhances wMOTA by 17.9\% and 7.2\%, respectively.

\begin{table}[h!]
    \centering
    \setlength{\tabcolsep}{2.4mm}
    \captionof{table}{\textbf{Evaluation on the nuScenes.}
    Results of behavior change-based Visual-ROI models on the nuScenes.}
    \begin{tabular}
            { @{} l @{} c @{} c @{} c @{} | c @{} c @{}}
            \toprule
              \multirow{2}{*}{ \begin{tabular}{@{\;}c@{\;}}
              \end{tabular}} & 
             \multicolumn{3}{c}{Spatial Accuracy} & 
             \multicolumn{2}{c}{Temporal Consistency}
             \\
             \cmidrule(lr){2-4} \cmidrule(lr){5-6} &
             \begin{tabular}{c} OT-P $\uparrow$ \end{tabular} & 
             \begin{tabular}{c} OT-R $\uparrow$ \end{tabular} & 
             \begin{tabular}{c} OT-F1 $\uparrow$ \end{tabular} &
             \begin{tabular}{c} PIC $\downarrow$ \end{tabular} & 
             \begin{tabular}{c}  wMOTA $\uparrow$ \end{tabular}
             \\
             \midrule
             BP~\cite{li2020gcn}
             & 21.1 & 38.0 & 27.1
             & 19.0 & 58.3
             \\
             BCP~\cite{li2020make}
             & \textbf{50.8} & 50.6 & 50.7
             & 15.2 & 69.0
             \\
             BS+BCP
             & 39.2 & 56.2 & 46.2
             & 10.7 & 65.3
             \\
             PF+BCP
             & 45.5 & \textbf{73.2} & \textbf{56.1}
             & \textbf{8.9} & \textbf{76.2}
             \\
            \bottomrule
    \end{tabular}
    \label{table:nuscenes}
\end{table}

Finally, utilizing the potential field in place of BEV-SEG (\textbf{BS+BCP}) resulted in improvements of 9.9\%, 1.8\%, and 10.9\% in F1, PIC, and wMOTA, respectively.
These results follow the trends observed in the synthetic data, demonstrating that our proposed model generalizes effectively to real-world scenarios.
Notably, both \textbf{BS+BCP} and \textbf{PF+BCP} exhibit substantially inferior performance compared to \textbf{BCP} in terms of OT-P.
We conjecture that this decline in performance stems from the limitations of BEV-SEG predictions, particularly in complex real-world scenarios. 
Prediction errors likely result in the misclassification of risk objects, contributing to the observed degradation.

Fig.~\ref{fig:qualitative_ROI} presents the qualitative results across various traffic scenarios.
We observe that \textbf{BCP} exhibits several false alarms, while \textbf{PF+BCP} reliably identifies risk objects, such as crossing pedestrians. 
Obviously, the proposed method avoids misidentifying vehicles in the opposite lane as risk objects.
These results confirm that our method effectively identifies potential risks and significantly reduces false alarms in complex scenarios, ensuring reliability in real-world applications.
%

\begin{figure}[t!]
    \centering
    \includegraphics[width=0.9\columnwidth]{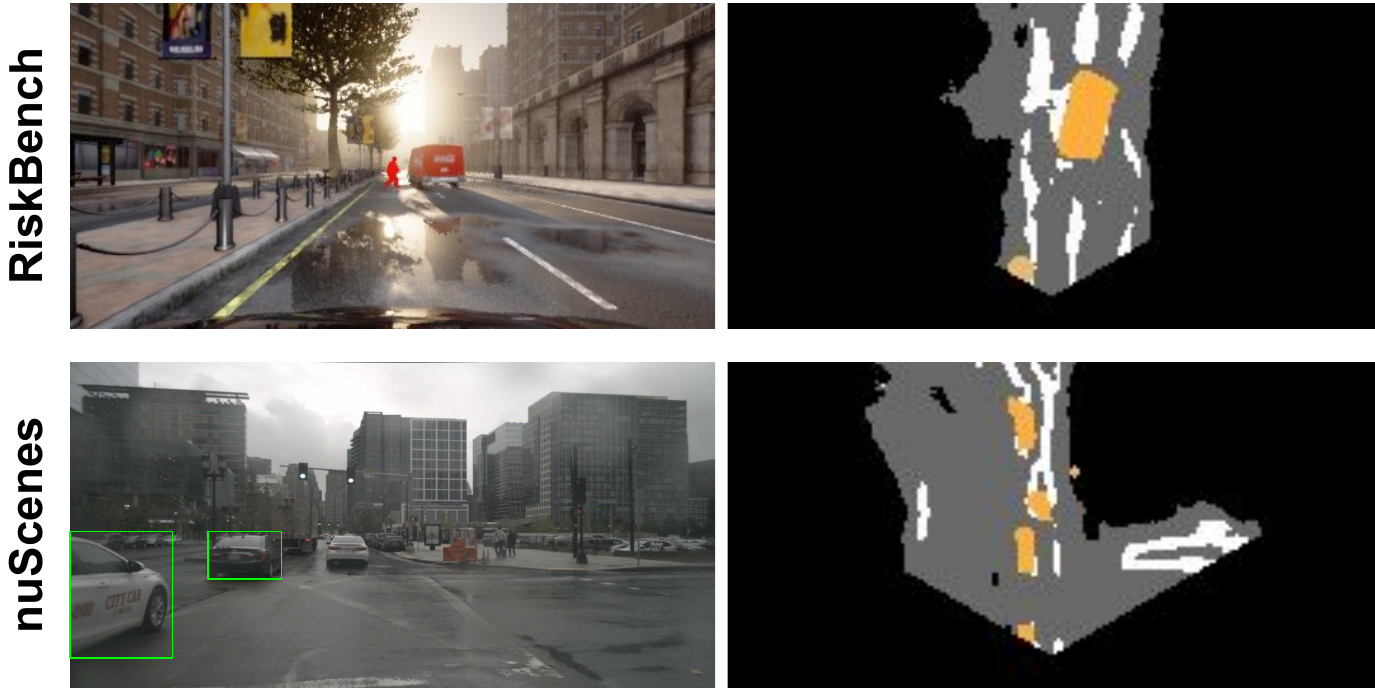}
    \caption{
    Failure cases from the \textbf{RiskBench} and \textbf{nuScenes} datasets.
    \textbf{Top:} A Pedestrian was too small to be detected by the perception model.
    \textbf{Bottom:} The absence of a roadline at the intersection resulted in a false positive due to the lack of clear road affordance.}
    \label{fig:qualitative_failure}
    \vspace{-3ex}
\end{figure}

%% file: Content/6-conclusion.tex
\label{sec:Conclusion}
In this work, we propose using potential fields as a new representation for scene affordance. 
We demonstrate that integrating potential fields can address the limitations of existing behavior change-based Visual-ROI algorithms, specifically, spatial inaccuracy, temporal inconsistency and computational inefficiency. 
Through comprehensive experiments and ablation studies on the RiskBench dataset, we achieve relative enhancements of 11.6\% in wMOTA and 20.3\% in the F1 score compared to strong baselines. Moreover, the proposed method can speed up causal inference by 88\%.
We also evaluate our approach on the nuScenes dataset, achieving a 5.4\% improvement in wMOTA and a 7.2\% improvement in the F1 score. 
%
%
These results underscore the vital role of scene affordance in advancing the safety and efficiency of intelligent driving systems.

\noindent\textbf{Limitation and Future Work:} 
The effectiveness of the potential fields in our approach is highly dependent on the quality of BEV semantic segmentation. 
For instance, Fig.~\ref{fig:qualitative_failure} shows failure cases due to imperfect BEV segmentation.
Manually defining repulsive and attractive force constants limits the broader applicability of our method.
To address this, we plan to explore alternative potential field rendering strategies, such as the approach in~\cite{nmp_zeng_2019}, and investigate more diverse scene affordances, like drivable areas and lane concepts, for integration into decision-making tasks.

\noindent\textbf{Acknowledgement:}
The work is sponsored in part by the Higher Education Sprout Project of the National Yang Ming Chiao Tung University and Ministry of Education (MOE), the Yushan Fellow Program Administrative Support Grant, and the National Science and Technology Council (NSTC) under grants 112-2634-F-002-006, 113-2634-F-002-007, 113-2628-E-A49-022, and Institute for Information Industry (III).